\documentclass[letterpaper]{article} % DO NOT CHANGE THIS
\usepackage{aaai2026}  % DO NOT CHANGE THIS
\usepackage{times}  % DO NOT CHANGE THIS
\usepackage{helvet}  % DO NOT CHANGE THIS
\usepackage{courier}  % DO NOT CHANGE THIS
\usepackage[hyphens]{url}  % DO NOT CHANGE THIS
\usepackage{graphicx} % DO NOT CHANGE THIS
\usepackage{natbib}  % DO NOT CHANGE THIS AND DO NOT ADD ANY OPTIONS TO IT
\usepackage{caption} % DO NOT CHANGE THIS AND DO NOT ADD ANY OPTIONS TO IT
\usepackage{algorithm}
\usepackage{algorithmic}
\usepackage{makecell}
\usepackage{multirow}
\usepackage{amsmath}
\usepackage{subcaption}
\usepackage{amssymb}
\usepackage{booktabs}
\usepackage{xcolor}         % colors
\usepackage{colortbl}
\usepackage{color}

\usepackage{newfloat}
\usepackage{listings}
\DeclareCaptionStyle{ruled}{labelfont=normalfont,labelsep=colon,strut=off} % DO NOT CHANGE THIS
\floatstyle{ruled}
\newfloat{listing}{tb}{lst}{}
\floatname{listing}{Listing}
\title{PatchTraj: Unified Time-Frequency Representation Learning via Dynamic Patches for Trajectory Prediction}
\author {
    Yanghong Liu\textsuperscript{\rm 1},
    Xingping Dong\textsuperscript{\rm 1}\footnotemark[1],
    Ming Li\textsuperscript{\rm 1},
    Weixing Zhang\textsuperscript{\rm 2},
    Yidong Lou\textsuperscript{\rm 2}
}
\affiliations {
    \textsuperscript{\rm 1}School of Computer Science, Wuhan University\\
    \textsuperscript{\rm 2}GNSS Research Center, Wuhan University\\
    \{yanghongliu, xingpingdong, liming751218, zhangweixing89, ydlou\}@whu.edu.cn
}

\begin{document}

\maketitle
% \renewcommand{\thefootnote}{\fnsymbol{footnote}}
% \footnotetext[1]{Corresponding author.}

\begin{abstract}
Pedestrian trajectory prediction is crucial for autonomous driving and robotics. While existing point-based and grid-based methods expose two main limitations: insufficiently modeling human motion dynamics, as they fail to balance local motion details with long-range spatiotemporal dependencies, and the time representations lack interaction with their frequency components in jointly modeling trajectory sequences. To address these challenges, we propose PatchTraj, a dynamic patch-based framework that integrates time-frequency joint modeling for trajectory prediction. Specifically, we decompose the trajectory into raw time sequences and frequency components, and employ dynamic patch partitioning to perform multi-scale segmentation, capturing hierarchical motion patterns. Each patch undergoes adaptive embedding with scale-aware feature extraction, followed by hierarchical feature aggregation to model both fine-grained and long-range dependencies. The outputs of the two branches are further enhanced via cross-modal attention, facilitating complementary fusion of temporal and spectral cues. The resulting enhanced embeddings exhibit strong expressive power, enabling accurate predictions even when using a vanilla Transformer architecture. Extensive experiments on ETH-UCY, SDD, NBA, and JRDB datasets demonstrate that our method achieves state-of-the-art performance. Notably, on the egocentric JRDB dataset, PatchTraj attains significant relative improvements of 26.7\% in ADE and 17.4\% in FDE, underscoring its substantial potential in embodied intelligence.
\end{abstract}

% Uncomment the following to link to your code, datasets, an extended version or similar.
% You must keep this block between (not within) the abstract and the main body of the paper.
% \begin{links}
%     \link{Code}{https://aaai.org/example/code}
%     \link{Datasets}{https://aaai.org/example/datasets}
%     \link{Extended version}{https://aaai.org/example/extended-version}
% \end{links}

\section{Introduction}
Pedestrian trajectory prediction is critical for autonomous driving~\cite{book1,book2,book65} and robotics~\cite{book3,book4}, where accurately forecasting future paths from observed trajectories hinges on effectively modeling spatiotemporal dependencies. Existing methods predominantly perform temporal-domain modeling of pedestrian trajectories, leveraging their natural capacity to capture dynamic motion evolution, such as position shifts and velocity changes, through sequential models like LSTMs~\cite{book17} or Transformers~\cite{book34}. These approaches excel at modeling local continuity and short-term motion trends, forming the backbone of most state-of-the-art systems.

However, time-domain modeling alone overlooks a crucial aspect: pedestrian motion also exhibits strong regularity in the frequency domain, as revealed by recent studies~\cite{book18,book19}. Frequency components compactly encode periodic patterns (e.g., gait cycles) and energy distributions, filtering noise while highlighting long-range dependencies that are often obscured in raw time-series data~\cite{book20}. Despite these advantages, joint modeling of time and frequency domains remains underexplored.
%, with a few works attempting to unify their complementary strengths.

\begin{figure}[!t]
	\centering
		\includegraphics[width=0.46\textwidth]{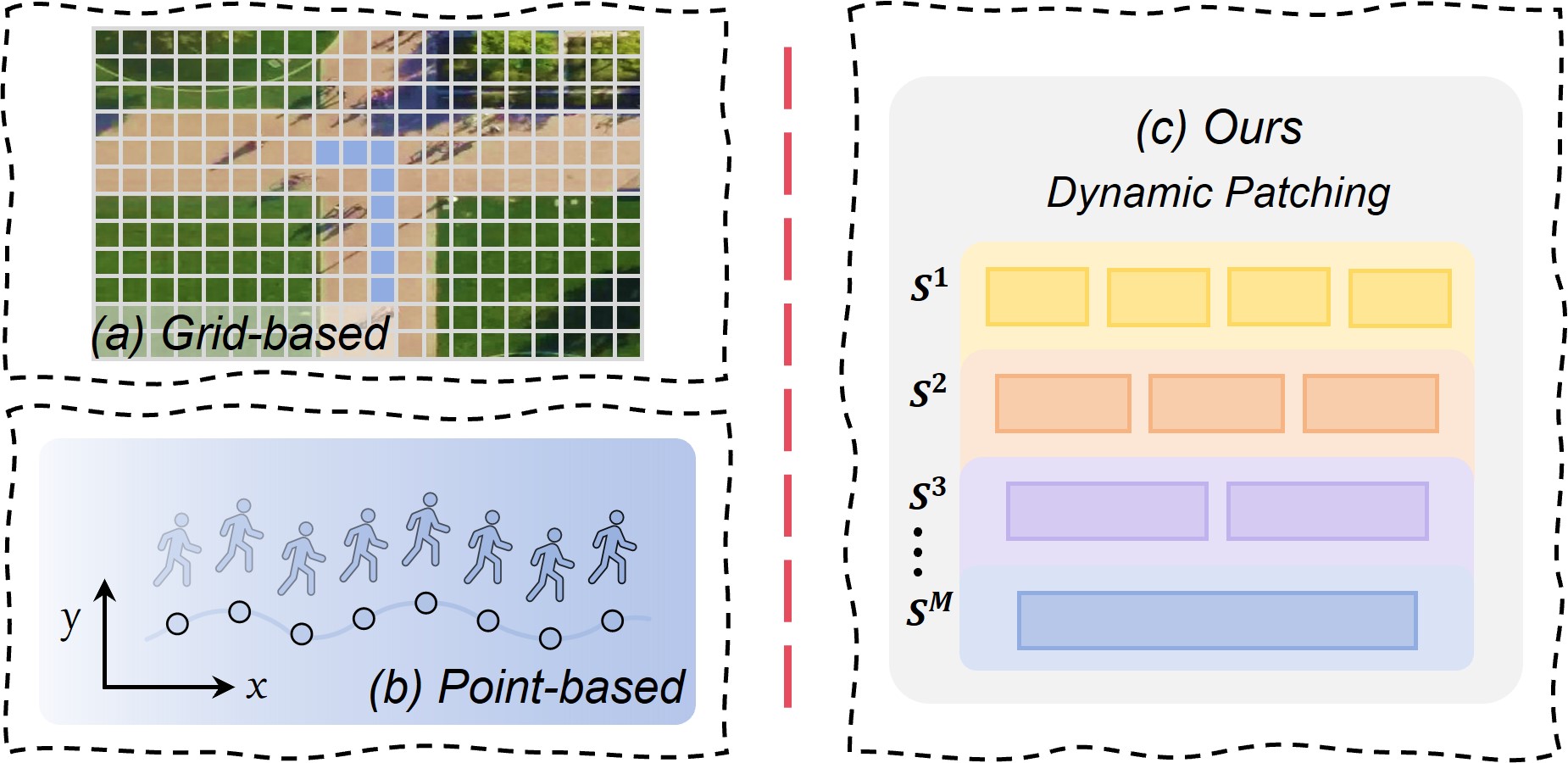}
	% \captionsetup{font={scriptsize}}
	\caption{Comparison between existing (a) grid-based methods, and (b) point-based methods for multi-modal trajectory prediction. The former is limited by fixed grid resolution, while the latter loses holistic motion semantics. Our method (c) introduces a dynamic patch mechanism to capture hierarchical local dynamics and comprehensive global semantics.
 }\label{fig:1}
 % \vspace{-4mm}
\end{figure}

Motivated by this gap, we initially experimented with a straightforward fusion approach within a dual-branch Transformer framework. Time-domain trajectories and their frequency projections were separately embedded and combined through concatenation. The fused features then served as input tokens of Transformer to predict future trajectory. While this yielded modest improvements, the gains were inconsistent, failing to fully exploit cross-domain synergy. We observe that the fundamental limitation lies in the trajectory representation itself. As illustrated in Figure~\ref{fig:1}, existing methods predominantly adopt either grid-based representation~\cite{book7,book8,book9} or point-based representation~\cite{book66,book6,book10}. Figure~\ref{fig:1} (a) divides the environment into fixed cells, leading to loss of precision in representing continuous trajectories. 
Figure~\ref{fig:1} (b) treats trajectory as discrete sequences of coordinates, losing holistic semantics.
Neither paradigm can simultaneously capture hierarchical local dynamics and comprehensive global semantics, thereby limiting the predictive capability of subsequent trajectory models. This motivates our rethinking of trajectory representation as structured, semantically meaningful patches.

Drawing inspiration from patch-based representations, which have demonstrated unique advantages in unifying local and global features in computer vision~\cite{book11,book12,book13} and time-series analysis~\cite{book14,book15,book16}. By adaptively segmenting trajectories into multi-scale spatiotemporal patches where each represents a semantically cohesive motion segment (e.g., a "stepping stride" or "waiting pause"), we enable hierarchical feature learning that preserves local details while modeling long-range dependencies.

In this paper, we present \textbf{PatchTraj}, a novel dynamic patch-based trajectory prediction framework that unifies time-frequency modeling through dynamic spatiotemporal patches. The framework first decomposes input trajectories into raw time-domain sequences and low-frequency components via Discrete Cosine Transform (DCT), preserving motion trends while filtering high-frequency noise. Instead of fixed-length segments, we introduce dynamic patch mechanism, where lightweight meta-network learns to group trajectory points into semantically meaningful patches based on motion dynamics. Each patch is processed by a Mixture-of-Experts (MoE)-enhanced embedding layer, followed by hierarchical feature aggregation through a Feature Pyramid Network (FPN) that fuses fine-grained and coarse-grained motion features. Additionally, the two branches interact via cross-modal attention, where time-domain queries attend to frequency-domain keys/values to enhance motion semantics. The unified representation is then processed by a Transformer encoder-decoder for autoregressive future trajectory prediction.
To summarize, the main contributions of this paper are as follows:
\begin{itemize}
    \item We propose a novel dual-branch Transformer-based framework, termed PatchTraj, with time-frequency hybridization for noise-robust trajectory modeling.
    \item We introduce the first dynamic patch mechanism to adaptively segment trajectory into variable-scale patches to capture multi-granularity motion patterns.
    \item We further present the scale-aware feature extraction strategy within PatchTraj. The multi-scale patches are processed by adaptive embedding layers, followed by hierarchical aggregator to model both fine-grained and long-range dependencies.
    \item We experimentally demonstrate that PatchTraj significantly outperforms previous state-of-the-art methods on four real-world datasets, including ETH-UCY, SDD, NBA, and JRDB.
\end{itemize}

\section{Related Work}
\subsection{Trajectory prediction}
Pedestrian trajectory prediction has become increasingly important for autonomous driving, surveillance, and robotics applications. Early approaches using physical models~\cite{book21} and traditional machine learning methods~\cite{book22,book23} established foundations but failed to capture real-world behavioral complexity. The field transformed with deep learning, particularly through RNN-based approaches ~\cite{book24} and LSTM architectures~\cite{book10} that effectively modeled temporal patterns. Subsequent advances introduced graph-based methods~\cite{book25,book26} to better represent group dynamics and long-range interactions. To handle behavioral uncertainty, researchers developed probabilistic techniques including CVAE~\cite{book27,book28} and GAN-based approaches~\cite{book29,book30} for multi-modal trajectory generation. Most recently, diffusion models~\cite{book31,book32,book33} have emerged as a promising direction, generating trajectories through iterative denoising processes.

\begin{figure*}[!t]
	\centering
		\includegraphics[width=0.94\textwidth]{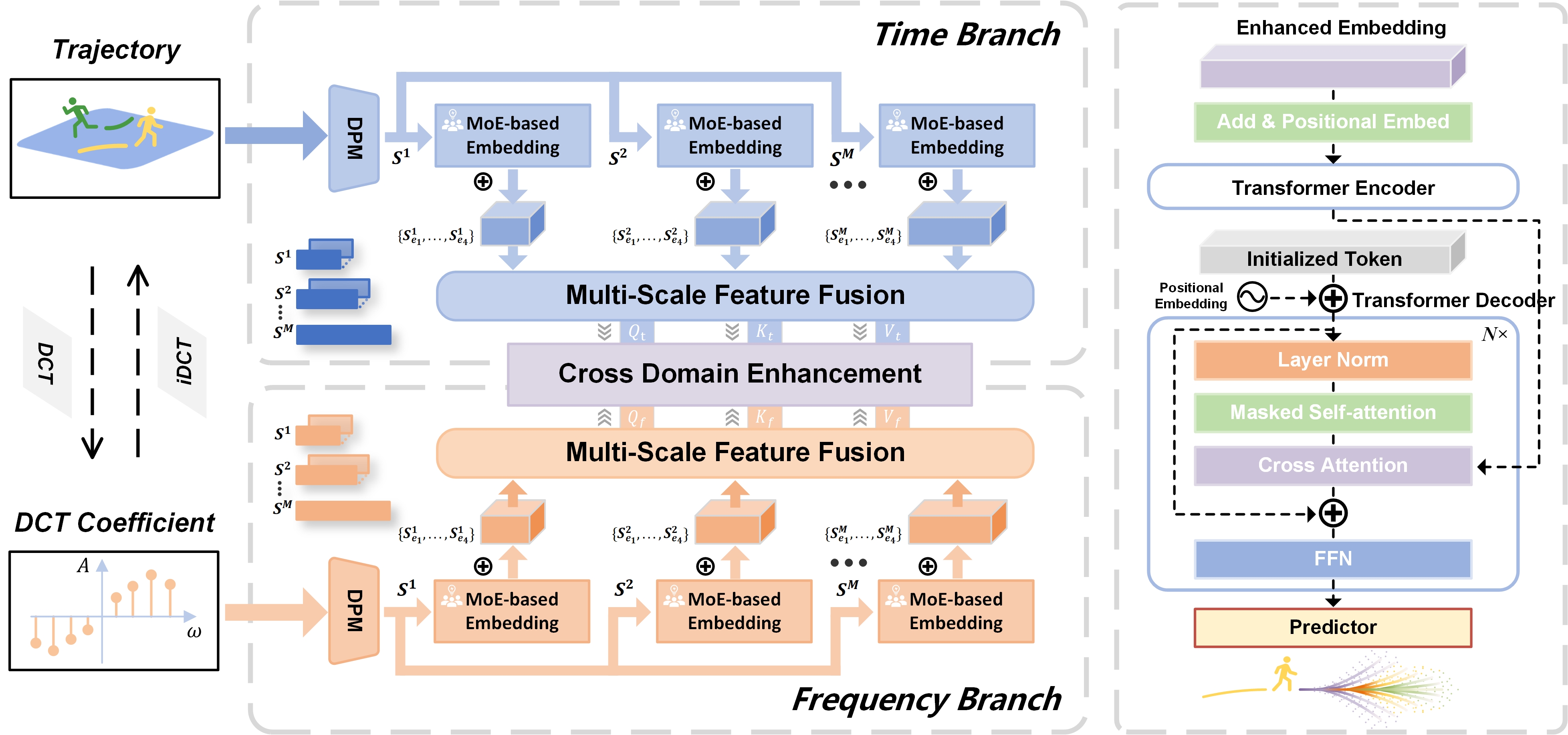}
	%\captionsetup{font={scriptsize}}
	\caption{The overall architecture of \textbf{PatchTraj}, which is a dual-branch trajectory prediction framework integrating time-domain and frequency-domain processing. The raw time sequences can be transferred into spectral components via DCT/iDCT. The dynamic patch mechanism is designed to capture multi-granularity motion patterns. Each patch is then processed by MoE-based embedding layer, and a FPN architecture is utilized to hierarchically aggregate multi-scale features. Finally, a Transformer encoder-decoder fuses both modalities and predicts future trajectories.}\label{fig:2}
    % \vspace{-2mm}
\end{figure*}

Transformers have revolutionized trajectory prediction by overcoming the limitations of sequential models in modeling long-range dependencies and complex interactions. Originally successful in NLP~\cite{book34}, they were adapted for motion forecasting through self-attention mechanisms that explicitly capture agent relationships and temporal dynamics. Pioneering works like Trajectron++~\cite{book6} combined graph networks with Transformer decoders to generate socially compliant trajectories, while AgentFormer~\cite{book35} employed a spatio-temporal attention to jointly encode agent histories and future interactions. MART~\cite{book37} is proposed to effectively capture individual and group behaviors via relational transformer mechanisms~\cite{book36}.

While most trajectory prediction methods operate solely in the time domain, recent work has revealed valuable periodic patterns in the frequency domain. Several approaches have explored this direction: V$^{2}$Net~\cite{book18} employs Fourier transforms for hierarchical frequency decomposition, DiffWT~\cite{book38} integrates wavelet transforms with diffusion models, and SpectrumNet~\cite{book39} encodes frequency features for RNN prediction. However, these methods face critical limitations with static frequency decomposition, complex wavelet processing, and deterministic outputs, preventing them from fully exploiting the complementary relationship between time and frequency domains for optimal prediction performance.

\subsection{Patch in time series forecasting}
Recent advances in time series forecasting have adopted patch-based modeling, inspired by the success of vision transformers (ViTs)~\cite{book13} in computer vision, where images are split into non-overlapping patches for efficient self-attention computation. This approach addresses key limitations of RNNs and CNNs, such as limited receptive fields and high computational cost for long sequences. Unlike traditional methods that process time series point-wise or sliding-window-based (e.g., ARIMA, LSTMs), patch-based approaches segment time series into localized temporal patches, enabling efficient long-range dependency modeling and hierarchical feature learning.

The PatchTST model~\cite{book15} pioneered the use of fixed-length patches in time series forecasting, demonstrating improved accuracy through local trend capture and efficient attention computation.
 Subsequent work has expanded this paradigm in several directions: TimesNet~\cite{book40} employs FFT-based multi-scale patching to identify periodic patterns across resolutions, while FiLM~\cite{book43} incorporates patch-level contrastive learning for cross-dataset generalization. To address computational complexity, methods like Informer~\cite{book42} and FEDformer~\cite{book41} combine patching with sparse or frequency-optimized attention mechanisms. PatchMLP~\cite{book16} utilize the MLPs to model interactions among cross variable, which achieves superior performance over existing Transformer models. While these advances establish patch-based modeling as effective for general time series analysis, their potential for trajectory prediction, particularly in capturing the spatiotemporal dynamics of pedestrian motion, remains unexplored.

\section{Method}
% We begin by outlining the problem setup and providing a concise overview of our framework architecture as illustrated in Figure~\ref{fig:2}.

\subsection{Problem Formulation}
Formally, the problem of pedestrian trajectory prediction can be formulated in the temporal domain as follows. For a given pedestrian $i$ in a scene, the observed trajectory over a historical time horizon $t=1,...,T_{obs}$ is represented as: $\mathbf{X}^{i}=[\mathbf{x}^{i}_{1},\mathbf{x}^{i}_{2},...,\mathbf{x}^{i}_{T_{obs}}] \in \mathbb{R}^{T_{obs} \times d}$, where $\mathbf{x}_{t}^{i}=(x^{i}_{t}, y^{i}_{t})$ denotes the 2D spatial coordinates (or $d$-dimensional state, e.g., including velocity, acceleration) at time $t$. For multi-agent settings, the input includes trajectories of all pedestrians: $\mathbf{X}=\{ \mathbf{X}^{1}, \mathbf{X}^{2},...,\mathbf{X}^{N}\}$, where $N$ is the number of pedestrians. The goal is to predict the future trajectory distribution over a prediction horizon $t=T_{obs}+1,...,T_{pred}$: $p(\mathbf{Y}^{i} | \mathbf{X}^{i}, \mathbf{X}^{-i}, \mathbf{C})$, where $\mathbf{Y}^{i}=[\mathbf{y}^{i}_{T_{obs}+1},..., \mathbf{y}^{i}_{T_{pred}}] \in \mathbb{R}^{T_{pred} \times d}$ is the future trajectory. $\mathbf{X}^{-i}$ denotes the observed trajectories of other pedestrians (for social interaction modeling) and $\mathbf{C}$ represents contextual information.

We also formulate the frequency-domain transformation of time trajectories using the Discrete Cosine Transform (DCT)~\cite{book53} to enable joint time-frequency analysis in our dual-branch framework shown in Figure~\ref{fig:2}. For a given observed trajectory $\mathbf{X}^{i}$, we firstly padding the past trajectory $T_{pred}$ times with the last timestep to form a $T$ length sequence, where $T = T_{obs}+T_{pred}$. Then apply Type-II DCT along the temporal axis to obtain spectral coefficients $\mathbf{c}^{i} \in \mathbb{R}^{T \times d}$: $\mathbf{c}^{i}_{n}=\operatorname{DCT}(\mathbf{x}_{t}^{i})=\sum\nolimits_{t=1}^{T} \sqrt{\frac{2}{N}} \mathbf{x}_{t}^{i} \cos[\frac{\pi}{2T}(2t-1)(n-1)]$, where $n \in {1,...,T}$ indexes frequency components. We then truncate the spectrum to retain the most informative $l \ll T$ coefficients: $\mathbf{c}^{i}=\mathbf{c}^{i}_{1:l}\in \mathbb{R}^{l \times d}$. Due to the DCT orthogonal property, the original data can be reconstructed via iDCT operation. 

% \subsection{Dynamic Patch Embedding}
\subsection{Dynamic Patch Mechanism}
Effective time series modeling requires precise detection of localized patterns and efficient model refinement. Segmenting data into temporal windows offers the model focused snapshots of short-term series behavior, improving its ability to characterize localized patterns and process these detailed features with greater accuracy. Conventional approaches are used to employ static patches for time series embedding, resulting in models that primarily recognize temporal patterns at a fixed scale, overlooking the inherent multi-scale dynamics and intricate relationships present in temporal data.

To capture local information in more detail and to fully understand the spital-temporal relationships within the observed trajectory, we propose the dynamic patch mechanism (DPM). Compared to static patching, which divides sequences into fixed-length segments, we define a collection of $M$ patch size values as $\mathcal{S}={S_{1},...,S_{M}}$, adaptively adjusting the patch size based on the different trajectory length. For a particular patch size $S\in\mathcal{S}$, the observed trajectory $\mathbf{X}$ is first divided into non-overlapping $P$ patches ($P=T/S$). Note that it is necessary to ensure the patch size $S$ is divisible by the time length $T$. After $M$ times patch division, we obtain a multi-scale patch set with various sizes $(\mathbf{X}_{P_{1}},...,\mathbf{X}_{P_{M}})$. The dynamic patch mechanism promotes the spital-temporal learning ability of trajectory with multi-scale representation, adaptively capturing hierarchical local dynamics and comprehensive global semantics.

\subsection{Multi-Scale Patch Embedding}
To effectively process the multi-scale patch representations generated by our dynamic patch mechanism, we propose a Mixture-of-Experts~\cite{book64} based embedding architecture (MSPE). This design enables specialized processing of different temporal granularities while maintaining computational efficiency through sparse expert activation.

Specifically, each expert in our MSPE module is designed to handle specific temporal scales, with dedicated projection weights and positional encodings for every patch size $S \in \mathcal{S}$. This specialization allows individual experts to develop optimized feature extraction capabilities for their assigned temporal granularities. A learnable gating network analyzes the global trajectory context to compute routing weights:
\begin{equation}
\mathbf{G} = \text{Softmax}(\text{MLP}(\text{Flatten}(\mathbf{X}))),
\end{equation}
where $\mathbf{G} \in \mathbb{R}^{B \times N \times M}$ contains the gating weights for $B$ sequences, $N$ experts, and $M$ patch scales.

For each input sequence, only top-$k$ experts are activated per patch scale, implementing conditional computation:
\begin{equation}
\mathbf{U}_{m} = \sum\nolimits_{n=1}^N \mathbb{I}(n \in \text{TopK}(\mathbf{G}{:,n,m})) \cdot \mathbf{W}_n^m(\mathbf{X}_{P_m}),
\end{equation}
where $\mathbf{W}_n^m$ denotes expert $n$ processing patches of size $S_m$, $\mathbb{I}(\cdot)$ uses 0 or 1 to indicate whether expert $n$ is selected. Due to the expert performance at different scales, we aggregate each expert's learned features along the different scale to construct a comprehensive multi-scale representation.

Our MSPE module elegantly handles multi-scale trajectory patches through a combination of dynamic gating and specialized expert processing. The system intelligently routes different patch scales to appropriate experts using learned attention weights, while maintaining scale-specific projections and positional encodings to preserve unique characteristics at each granularity level.

% The outputs from different scales are combined through a learnable aggregation layer:
% \begin{equation}
% \mathbf{Z} = \sum\nolimits_{m=1}^M \alpha_m \cdot \mathbf{U}_m
% \end{equation}
% with $\alpha_m$ being the importance weights automatically learned for each temporal scale.

\subsection{Multi-Scale Feature Fusion}
The multi-scale feature fusion (MSFF) module aggregates dynamic patch representations from both temporal and frequency domains through a Feature Pyramid Network architecture. First, expert-processed features at each patch scale are averaged to generate scale-specific embeddings, which are then reorganized into a spatial format compatible with 1D convolutions. 
For each patch size $p_s \in \mathcal{P}$ where $\mathcal{P}$ is the set of patch scales, we aggregate all expert features:
\begin{equation}
\mathbf{\bar{U}}_{p_s} = \frac{1}{|\mathcal{E}_{p_s}|}\sum_{n\in\mathcal{E}_{p_s}}\mathbf{U}_{p_s}^n,
\end{equation}
where $\mathcal{E}_{p_s}$ represents the set of experts processing scale $p_s$, and $\mathbf{U}_{p_s}^n$ is the feature from expert $n$ at scale $p_s$.

 The MSFF module processes diverse scales in a descending order ($p_1 > p_2 > \cdots > p_{M}$), where higher-resolution features are progressively downsampled and combined with coarser-scale representations through lateral connections. For scale $p_i$, the lateral connection transforms the features:
\begin{equation}
\mathbf{F}_{p_i} = \text{Conv1D}(\mathbf{\bar{U}}_{p_i}).
\end{equation}

The top-down pathway combines features through recursive upsampling:
\begin{equation}
\mathbf{P}_{p_i}  = \begin{cases}
\text{Conv1D}(\mathbf{F}_{p_i}), & \text{if } i=1, \\
\text{Conv1D}(\mathbf{F}_{p_i} + \text{Upsample}(\mathbf{P}_{p_{i-1}})), & \text{otherwise},
\end{cases}
\end{equation}
features are further refined through top-down propagation:
\begin{equation}
\mathbf{E}_{p_i} = \begin{cases}
\mathbf{P}_{p_n}, & \text{if } i=n, \\
\mathbf{P}_{p_i} + \text{Upsample}(\mathbf{E}_{p_{i+1}}), & \text{otherwise},
\end{cases}
\end{equation}
where $\mathbf{E}_{p_i}$ represents the enhanced feature at scale $p_i$. 

To ensure dimensional consistency during fusion, adaptive padding or truncation is applied when merging features across scales. The final outputs with each branch formed as $\mathcal{F}_{t}$ and $\mathcal{F}_{f}$, which select the most refined scale after enrichment from all coarser levels, producing a unified representation that preserves high-resolution details while integrating multi-scale motion patterns. This hierarchical fusion enables the model to simultaneously capture local trajectory dynamics and global temporal structures, essential for robust trajectory prediction in complex scenarios.

\subsection{Cross Domain Enhancement}
While Figure~\ref{fig:2} illustrates our dual-branch architecture for processing time and frequency inputs, we further bridge these modalities by interacting final fused features $\mathcal{F}_{t}$ and $\mathcal{F}_{f}$ through the cross-attention mechanism. To compensate for the positional insensitivity of patch-based representations, we augment both feature streams with learnable positional encodings. The enhancement module then establishes bidirectional interaction via:

\textbf{Temporal-to-frequency attention} enables time features to dynamically attend to relevant frequency components. The temporal patches serve as queries to retrieve complementary spectral information through dot-product attention:
\begin{equation}
\mathcal{\hat{F}}_{t}=\text{Attention}(Q_t,K_f,V_f) = \text{softmax}(\frac{Q_t K_f^T}{\sqrt{d_k}})V_f,
\end{equation}
where $Q_t$ denotes time-patch embeddings $\mathcal{F}_{t}$ and $K_f,V_f$ represent frequency-domain keys/values from $\mathcal{F}_{f}$.

\textbf{Frequency-to-temporal attention} conversely allows spectral features to assimilate critical temporal patterns. This reverse attention flow helps localize periodic motion characteristics in time domain.
Residual connection preserves original modality-specific features while incorporating cross-domain enhancements through element-wise addition:
\begin{equation}
    \mathcal{F}_t' = \mathcal{F}_t + \mathcal{\hat{F}}_{t}, \quad \mathcal{F}_f' = \mathcal{F}_f + \mathcal{\hat{F}}_{f}.
\end{equation}

The concatenated $[\mathcal{F}_t'; \mathcal{F}_f']$ maintain both domains' distinctive properties while capturing their latent correlations. 

\subsection{Transformer Encoder}
Our framework employs a vanilla Transformer encoder comprising $N$ stacked blocks with skip connections~\cite{book44} to process fused multi-modal patch embeddings. The raw time-domain trajectories are first encoded via MLPs, with modality-aware features preserved through residual connections to maintain temporal awareness. Within each transformer block, multi-head self-attention operates on the patch embeddings, stabilized by layer normalization and skip connections.
\begin{table*}[!ht]
        \scriptsize
	\centering
     \caption{Quantitative comparison results on (a) JRDB, (b) NBA, (c) SDD and (d) ETH-UCY datasets. \textbf{ADE} and \textbf{FDE} are reported on (a) for deterministic prediction, \textbf{minADE$_{20}$} and \textbf{minFDE$_{20}$} are reported on (b), (c), (d) for multi-modal prediction. \textbf{Bold} and \underline{underlined} fonts represent the best and second-best results, respectively (lower values are better).}\label{table1}
	\renewcommand\tabcolsep{1.6pt}
	\renewcommand{\arraystretch}{1.0}
	\begin{tabular}{lccccccc>{\columncolor{gray!10}}c}
        % \begin{tabular}{lcccccccc}
		%\Xhline{1pt} 
        \toprule
        \multicolumn{9}{c}{\textbf{(a) JRDB Dataset} ($K=1$)} \\ 
        \midrule 
		\multirow{2}{*}{Time} & Social-GAN  & Trajectron++ & EqMotion & LED  &  Social-Trans & EmLoco & NMRF &  \\ 
        &  \cite{book29}  & \cite{book6} & \cite{book55} & \cite{book32}  &  \cite{book56} & \cite{book63} & \cite{book57} & \multirow{-2}{*}{Ours} \\
        \midrule  
         4.8s & 0.50/0.99 & 0.40/0.78 & 0.42/0.78 & 0.32/0.54 & 0.40/0.77 & 0.37/0.72 & \underline{0.26}/\underline{0.48} &  \textbf{0.20}/\textbf{0.40}  \\ 
	\toprule
        \toprule
        \multicolumn{9}{c}{\textbf{(b) NBA Dataset} ($K=20$)} \\ 
        \midrule 
		\multirow{2}{*}{Time} & STAR & Trajectron++ & MemoNet &  GroupNet & LED & MART  & NMRF &   \\
          & \cite{book58} & \cite{book6} & \cite{book59}  &  \cite{book28} & \cite{book32} & \cite{book37}  & \cite{book57} & \multirow{-2}{*}{Ours}   \\
        \midrule  
         4.0s & 1.13/2.01 & 1.15/1.57 & 1.25/1.47 & 0.96/1.30 & 0.81/1.10 & \underline{0.73}/\textbf{0.90} & 0.75/0.97  & \textbf{0.68}/\underline{0.94} \\ 
	\toprule	
        \toprule
        \multicolumn{9}{c}{\textbf{(c) SDD Dataset} ($K=20$)} \\ 
        \midrule 
		\multirow{2}{*}{Time} &  V$^{2}$Net & GroupNet & MemoNet & LED  & MGF & MART & NMRF &   \\ 
         &  \cite{book18} & \cite{book28} & \cite{book59} & \cite{book32}  & \cite{book61} & \cite{book37} & \cite{book57} & \multirow{-2}{*}{Ours}  \\
        \midrule  
         4.8s & \underline{7.12}/11.39 & 9.31/16.11 & 8.56/12.66 & 8.48/11.66 & 7.74/12.07 & 7.43/11.82 & 7.20/\underline{11.29}  & \textbf{6.58}/\textbf{11.14} \\ 
	\toprule	
        \toprule
        \multicolumn{9}{c}{\textbf{(d) ETH-UCY Dataset} ($K=20$)} \\ 
        \midrule 
		\multirow{2}{*}{Subset} & GroupNet &  AgentFormer & MemoNet & LED & MART & MoFlow  & NMRF &  \\ 
         & \cite{book28} &  \cite{book35} & \cite{book59} & \cite{book32} & \cite{book37} & \cite{book62}  & \cite{book57} & \multirow{-2}{*}{Ours}  \\
        \midrule  
        ETH  & 0.46/0.73 & 0.45/0.75 & 0.40/0.61 & 0.39/0.58  & 0.35/\underline{0.47} &  0.40/0.57 & \textbf{0.26}/\textbf{0.37}  & \underline{0.30}/0.48 \\ 
         HOTEL & 0.15/0.25 & 0.14/0.22 & \underline{0.11}/\underline{0.17} & \underline{0.11}/\underline{0.17}  & 0.14/0.22 & \underline{0.11}/\underline{0.17} & \underline{0.11}/\underline{0.17}  & \textbf{0.10}/\textbf{0.16} \\ 
         UNIV & 0.26/0.49 & 0.25/0.45 & \underline{0.24}/\underline{0.43} & 0.26/\underline{0.43}  & 0.25/0.45 & \textbf{0.23}/\textbf{0.39} & 0.28/0.49  & \textbf{0.23}/0.45 \\ 
       ZARA1 & 0.21/0.39 & 0.18/0.30 & 0.18/0.32 & 0.18/\textbf{0.26}  & 0.17/0.29 &  \underline{0.15}/\textbf{0.26} & 0.17/0.30  & \textbf{0.14}/\underline{0.27} \\ 
        ZARA2 & 0.17/0.33 & 0.14/0.24 & 0.14/0.24 & 0.13/\underline{0.22}  & 0.13/\underline{0.22} &  \underline{0.12}/\underline{0.22} &  0.14/0.25 & \textbf{0.10}/\textbf{0.19}  \\ \midrule
        AVG & 0.25/0.44 & 0.23/0.39 & 0.21/0.35 & 0.21/0.33  & 0.21/0.33 &  0.20/\underline{0.32} & \underline{0.19}/\underline{0.32}  &  \textbf{0.17}/\textbf{0.31} \\ 
	% \bottomrule
        \toprule
	\end{tabular}
    % \vspace{-4mm}
\end{table*}

\subsection{Trajectory Decoder}
The prediction pipeline employs an autoregressive decoder with $N$ transformer blocks to generate future trajectories from encoded representations. We initialize a learnable prediction token $\mathcal{T} \in \mathbb{R}^{T_{pred} \times D}$ as decoder inputs. Each decoder layer performs cross-attention between prediction queries and encoder output features. We use proper masking inside decoder to enforce causality of the decoder output sequence.
Final MLP head predicts trajectory coordinates:
\begin{equation}
\mathbf{\hat{Y}} = \text{MLP}(\mathcal{T}^{(N)} ) \in \mathbb{R}^{K \times T_{pred} \times 2},
\end{equation}
where output maintains multiple hypotheses ($K$ samples) for stochastic prediction.

\subsection{Training Constraint}
The framework is trained in an end-to-end strategy, we simultaneously optimize \textit{marginal loss}~\cite{book29} and \textit{joint loss}~\cite{book45} by employing L2-norm among $K$ samples to minimize the distance between the prediction and the ground truth. Please refer to our proof in supplementary materials for more details.
\begin{align}
    \mathcal{L}_{marginal} &= \sum\nolimits_{n}^{N} \operatorname{min}_{k}^{K}\|\mathbf{Y}_{n}-\hat{\mathbf{Y}}_{n}^{k} \|_{2}, \nonumber \\
    \mathcal{L}_{joint} &= \operatorname{min}_{k}^{K}\sum\nolimits_{n}^{N}\|\mathbf{Y}_{n}-\hat{\mathbf{Y}}_{n}^{k}\|_{2},
    \nonumber \\
    \mathcal{L} &= \mathcal{L}_{marginal} + \lambda\mathcal{L}_{joint}.
\end{align}

\section{Experiments}
\subsection{Datasets}
The model is trained and evaluated on four publicly available datasets for trajectory prediction: ETH-UCY~\cite{book46}, Stanford Drone Dataset (SDD)~\cite{book47}, NBA SportVU Dataset (NBA)~\cite{book48}, and the JackRabbot Dataset and Benchmark (JRDB)~\cite{book49}. \textbf{ETH-UCY} dataset comprises five distinct subsets: ETH, HOTEL, UNIV, ZARA1 and ZARA2. We follow the leave-one-out approach from~\cite{book29} with four subsets for training-validation and the remaining subset for testing. We utilize the standard  dataset splits to predict the future 12 frames (4.8s) with 8 frames observations (3.2s). \textbf{SDD} dataset captures large-scale pedestrian behaviors on a campus from a bird’s-eye perspective. The dataset is split into predicting the future 12 frames (4.8s) with 8 frames observations (3.2s). \textbf{NBA} dataset tracks the movements of 10 basketball players and a ball during NBA games in 2015-2016 season. The movements exhibit strong purpose fulness and complex variety, which increases difficulties compared to the pedestrian datasets. Followed by~\cite{book32}, we predict the future 20 frames (4.0s) based on 10 frames (2.0s) history. \textbf{JRDB} is a large-scale egocentric dataset recorded by a social robot in indoor and outdoor scenarios with stationary and moving behaviors. We follow the deterministic prediction split in~\cite{book56} and the multi-modal prediction split in~\cite{book57}, which predicts the future 12 frames (4.8s) over the observed 9 frames (3.6s).

\subsection{Implementation Details}
We represent each trajectory as a 6-dimensional vector ($d=6$) combining absolute position, relative displacement, and velocity.
For frequency-domain processing, we employ truncated DCT/iDCT operations retaining the first $l$ coefficients, with dataset-specific values: $l=8$ (ETH-UCY/SDD), $l=10$ (NBA), and $l=9$ (JRDB). This maintains dimensional consistency between time and frequency branches while preserving key spectral components. To obtain non-overlapping patches, we set a list of dynamic patch size $\mathcal{S}=\{2,4,8\}$ for ETH-UCY/SDD, $\mathcal{S}=\{2,5,10\}$ for NBA, $\mathcal{S}=\{1,3,9\}$ for JRDB according to the history length. Four experts are used in the MSPE module and top-2 experts are selected.
We employ $L = 4$ layers with $H = 4$ attention heads in each Transformer-based sub-network, and the dimension of hidden state and patch embedding is set $D=256$. The model is implemented in PyTorch~\cite{book50} and optimized using AdamW~\cite{book51} with a batch size of 12, an initial learning rate of $1\times10^{-3}$ which halve every 10 epochs, and trained for 200 epochs on a single 4090 GPU. 
For the loss coefficient weight, we set $\lambda=0.5$.

\subsection{Evaluation Metrics}
We adopt two widely reported metrics to evaluate the trajectory prediction performance: the Average Displacement Error (ADE) and the Final Displacement Error (FDE)~\cite{book52}. ADE and FDE measures the accuracy between the ground truth and predicted trajectory over all timestep and the last timestep. Inspired by~\cite{book29}, we adopt multi-modal prediction and generate $K$ samples for each trajectory. Consequently, minADE$_{K}$ and minFDE$_{K}$ are reported in our evaluation results.
% \begin{table}[!t]
%          \footnotesize
% 	\centering
% 	\caption{Multi-modal trajectory prediction comparisons on JRDB dataset. \textbf{minADE$_{20}$}/\textbf{minFDE$_{20}$} in meters are reported for the future 12 frames (4.8s). }\label{table2}
% 	\renewcommand\tabcolsep{6.0pt}
% 	% \renewcommand{\arraystretch}{1.2}
% 	\begin{tabular}{l|cccc}
% 		% \Xhline{1pt}
%         \toprule
% 		 Method & 1.2s & 2.4s & 3.6s & 4.8s   \\  \midrule
% 		LED~\cite{book32} & 0.05/0.07 & 0.09/0.14 & 0.14/0.21 & 0.18/0.28 \\  
%         NMRF~\cite{book57} & 0.04/0.05 &  0.08/0.11 & 0.11/0.17 & 0.15/0.23 \\ \midrule
% 		\rowcolor{gray!20}  Ours & \textbf{0.02}/$\textbf{0.03}$ & $\textbf{0.05}$/$\textbf{0.08}$ & $\textbf{0.08}$/$\textbf{0.13}$ & $\textbf{0.11}$/$\textbf{0.19}$ \\ 
% 	\bottomrule			
% 	\end{tabular}
%         \vspace{-4mm}
% \end{table}

\subsection{Quantitative Results}
\textit{JRDB.}
Table~\ref{table1}-(a) compares our method with 8 prominent methods on the JRDB dataset. For fair comparison with deterministic baselines, we configure PatchTraj to generate single-sample predictions. Our method establishes new state-of-the-art performance, demonstrating significant improvements of 23.1\% in ADE and 16.7\% in FDE over the previous best approach (NMRF). We also conduct experiments to compare the stochastic prediction, the superior performance presented in Table~\ref{table2} further investigate the effectiveness of our method.

\textit{NBA.}
Table~\ref{table1}-(b) compares our method with 8 prominent methods on the NBA dataset. Results show that our PatchTraj reduces ADE/FDE from 0.75/0.97 to 0.68/0.94 compared with NMRF in total 4 seconds, improving performance by 9.3\% and 3.1\%, respectively. We achieve the best performance on ADE, but the absence of explicit intention-aware modeling fundamentally limits FDE performance.

\textit{SDD.}
Table~\ref{table1}-(c) compares our method with 8 prominent methods on the SDD dataset, with a sample number limit of $K=20$ for all comparative methods. Our method maintains state-of-the-art accuracy to the best-of-20 samples. Specifically, we surpass outstanding spectral-based V$^{2}$Net by improving ADE/FDE from 7.12/11.39 to 6.58/11.14. Compared to the latest state-of-the-art method NMRF, PatchTraj reduces ADE from 7.20 to 6.58, achieving {8.6\%} large improvement. A reasonable interpretation is that we propose the dynamic patch mechanism to capture local motion patterns with trajectory segments.
\begin{figure*}[!t]
	\centering
		\includegraphics[width=\textwidth]{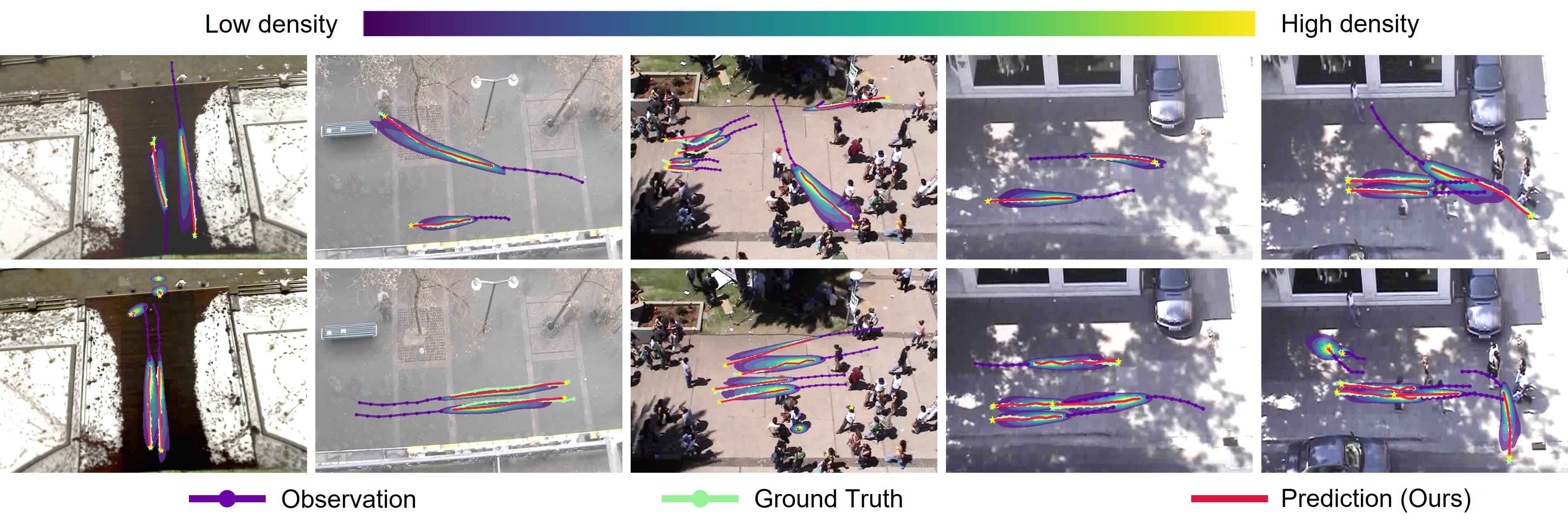}
	%\captionsetup{font={scriptsize}}
	\caption{Visualization results on the ETH-UCY dataset, including heatmaps of the overall distribution of predicted samples and best-of-20 predictions of our method.
    }\label{fig:3}
    % \vspace{-4mm}
\end{figure*}

\textit{ETH-UCY.}
As demonstrated in Table~\ref{table1}-(d), PatchTraj achieves superior performance compared to 8 state-of-the-art methods across nearly all ETH-UCY subsets. Our approach shows particular advantages over Transformer-based baselines (AgentFormer, MART) through its dynamic patch mechanism, which effectively captures multi-scale motion patterns. Notably, PatchTraj outperforms the current state-of-the-art NMRF by significant margins, reducing ADE by 10.5\% and FDE by 3.1\%. These consistent improvements across diverse scenarios validate the effectiveness of our method in pedestrian trajectory prediction.
\begin{table*}[!t]
	\centering
  	\caption{Ablations to study the contribution of key technical components on model performance. Average ADE and FDE are reported on NBA, SDD and ETH-UCY datasets. ``T'' denotes the time and ``F'' denotes the frequency. ``DPM'' indicates the dynamic patch mechanism, and ``MSPE'' means MoE-based multi-scale patch embedding. ``MSFF'' means multi-scale feature fusion with FPN, and ``CDE'' indicates the cross domain enhancement. $K$ is the number of multi-modal prediction.}
	\label{table3}
	\renewcommand\tabcolsep{3.0pt}
	\renewcommand{\arraystretch}{1.2}
	\footnotesize
	\begin{tabular}{cccccccccccccc}
		\hline\hline
		 \multirow{2}{*}{{vanilla Transformer}} & \multirow{2}{*}{{T-branch}} & \multirow{2}{*}{{F-branch}} & \multirow{2}{*}{{DPM}}  &  \multirow{2}{*}{{MSPE}} & \multirow{2}{*}{{MSFF}} & \multirow{2}{*}{{CDE}} & {N-Sample} & \multicolumn{2}{c}{{NBA}} & \multicolumn{2}{c}{{SDD}} & \multicolumn{2}{c}{{ETH-UCY}}  \\ 
        \cmidrule(lr){8-8}\cmidrule(lr){9-10}\cmidrule(lr){11-12}\cmidrule(lr){13-14}
		 &  &  &  &  &  &  & $K$ & ADE & FDE & ADE & FDE & ADE & FDE \\ \hline
		$\checkmark$ & $\checkmark$ & - & - & - & - & - &  20 & 1.02 & 1.43 & 8.03 & 15.30 & 0.32 & 0.55  \\ 
		$\checkmark$ & $\checkmark$ & $\checkmark$ & - & - & - & - &  20 & 0.96 & 1.38 & 7.68 & 14.24 & 0.30 & 0.52  \\ 
        $\checkmark$ & $\checkmark$ & $\checkmark$ & $\checkmark$ & - & - & - &  20 & 0.82 & 1.10 & 7.05 & 12.57 & 0.24 & 0.48  \\ 
		$\checkmark$ & $\checkmark$  & $\checkmark$ & $\checkmark$ & $\checkmark$ & - & - & 20 & 0.77 & 1.08 & 6.89 & 11.71 & 0.22 & 0.42  \\ 
        $\checkmark$ & $\checkmark$ & $\checkmark$ & $\checkmark$ & $\checkmark$ & $\checkmark$ & - &  20 & 0.71 & 1.00 & 6.68 & 11.64 & 0.19 & 0.36  \\ 
        $\checkmark$ & $\checkmark$ & $\checkmark$ & $\checkmark$ & $\checkmark$ & $\checkmark$ & $\checkmark$ &  20 & \textbf{0.68} & \textbf{0.94} & \textbf{6.58} & \textbf{11.14} & \textbf{0.17} & \textbf{0.31}  \\ 
		\hline\hline		
	\end{tabular}
    % \vspace{-4mm}
\end{table*}

\subsection{Qualitative Results}
To have a more intuitive awareness of our PatchTraj performance, we visualize the overall distribution and optimal outcome from 20 predictions in Figure~\ref{fig:3}. The results show that the predicted trajectories greatly align with the ground truth, demonstrating the effectiveness of our method. Please refer to the more visualizations in supplementary materials.

\subsection{Ablation Studies}
We further perform extensive ablation studies on NBA, SDD and ETH-UCY datasets illustrated in Table~\ref{table3} to investigate the contribution of key components in our method, with sample number fixed at $K=20$ for consistency. As shown in the first row of Table~\ref{table3}, the vanilla Transformer with separate time-domain inputs establishes our baseline performance. The second line suggests that constructing a dual-branch architecture to simultaneously integrate time and  frequency trajectory information can improve prediction performance. The third line demonstrates that our proposed dynamic patch mechanism enhances time-frequency representation through local dynamic and global dependency modeling. The two key components of ``MSPE'' and ``MSFF'' further improve model performance. The ``CDE'' module effectively establishes cross-modal relationships, enabling synergistic fusion of heterogeneous features. 
The cumulative improvements demonstrate that each component contributes uniquely to our state-of-the-art performance, with the complete framework achieving a large enhancement over the baseline. Please refer to the more ablations in supplementary materials.
\begin{table}[!t]
         \footnotesize
	\centering
	\caption{Multi-modal trajectory prediction comparisons on JRDB dataset. \textbf{minADE$_{20}$}/\textbf{minFDE$_{20}$} in meters are reported for the future 12 frames (4.8s). }\label{table2}
	\renewcommand\tabcolsep{6.0pt}
	\begin{tabular}{l|cc>{\columncolor{gray!10}}c}
        % \begin{tabular}{l|ccc}
		% \Xhline{1pt}
        \toprule
		 \multirow{2}{*}{Time} & LED & NMRF &  \\
          & \cite{book32} & \cite{book57} & \multirow{-2}{*}{Ours} \\  \midrule
		1.2s & 0.05/0.07 & 0.04/0.05 & \textbf{0.02}/\textbf{0.03}  \\  
        2.4s & 0.09/0.14 & 0.08/0.11 & \textbf{0.05}/\textbf{0.08}  \\ 
		3.6s & 0.14/0.21 & 0.11/0.17 & \textbf{0.08}/\textbf{0.13} \\
        4.8s & 0.18/0.28 & 0.15/0.23 & \textbf{0.11}/\textbf{0.19} \\ 
	\bottomrule			
	\end{tabular}
        % \vspace{-4mm}
\end{table}

\section{Conclusion}
In this paper, we present PatchTraj, a novel dynamic patch-based trajectory prediction framework that unifies time-domain and frequency-domain representations. Our approach addresses the limitations of existing methods by introducing a dynamic patch mechanism to capture multi-granularity motion patterns, a MoE-enhanced embedding layer for scale-aware feature extraction, and a hierarchical feature fusion module to balance fine-grained details with long-range dependencies. The cross-domain enhancement further leverages the complementary strengths of time and frequency representations, enabling robust and noise-resistant trajectory modeling. Extensive experiments on four real-world datasets (ETH-UCY, SDD, NBA, and JRDB) have demonstrated that PatchTraj achieves state-of-the-art performance, outperforming existing methods in both accuracy and robustness. Ablation studies validate the contributions of each key component, highlighting the effectiveness of our design choices.

\bibliography{references}

\begin{thebibliography}{63}
\providecommand{\natexlab}[1]{#1}

\bibitem[{Alahi et~al.(2016)Alahi, Goel, Ramanathan, Robicquet, Fei-Fei, and Savarese}]{book10}
Alahi, A.; Goel, K.; Ramanathan, V.; Robicquet, A.; Fei-Fei, L.; and Savarese, S. 2016.
\newblock Social lstm: Human trajectory prediction in crowded spaces.
\newblock In \emph{Proceedings of the IEEE conference on computer vision and pattern recognition}, 961--971.

\bibitem[{Berndt, Emmert, and Dietmayer(2008)}]{book23}
Berndt, H.; Emmert, J.; and Dietmayer, K. 2008.
\newblock Continuous driver intention recognition with hidden markov models.
\newblock In \emph{2008 IEEE Intelligent Transportation Systems Conference (ITSC)}, 1189--1194.

\bibitem[{Brown et~al.(2020)Brown, Mann, Ryder, Subbiah, Kaplan, Dhariwal, Neelakantan, Shyam, Sastry, Askell et~al.}]{book14}
Brown, T.; Mann, B.; Ryder, N.; Subbiah, M.; Kaplan, J.~D.; Dhariwal, P.; Neelakantan, A.; Shyam, P.; Sastry, G.; Askell, A.; et~al. 2020.
\newblock Language models are few-shot learners.
\newblock \emph{Advances in neural information processing systems}, 33: 1877--1901.

\bibitem[{Chai et~al.(2019)Chai, Sapp, Bansal, and Anguelov}]{book2}
Chai, Y.; Sapp, B.; Bansal, M.; and Anguelov, D. 2019.
\newblock Multipath: Multiple probabilistic anchor trajectory hypotheses for behavior prediction.
\newblock In \emph{Conference on Robot Learning}, 86--99.

\bibitem[{Chen et~al.(2024)Chen, Cao, Lin, Kitani, and Pang}]{book61}
Chen, J.; Cao, J.; Lin, D.; Kitani, K.; and Pang, J. 2024.
\newblock Mixed gaussian flow for diverse trajectory prediction.
\newblock In \emph{Advances in Neural Information Processing Systems}.

\bibitem[{Chen et~al.(2025)Chen, Zeng, Gao, Ding, and Bian}]{book38}
Chen, X.; Zeng, L.; Gao, M.; Ding, C.; and Bian, Y. 2025.
\newblock DiffWT: Diffusion-Based Pedestrian Trajectory Prediction With Time-Frequency Wavelet Transform.
\newblock \emph{IEEE Internet of Things Journal}, 12(5): 5109--5121.

\bibitem[{Diao and Loynd(2022)}]{book36}
Diao, C.; and Loynd, R. 2022.
\newblock Relational attention: Generalizing transformers for graph-structured tasks.
\newblock \emph{arXiv preprint arXiv:2210.05062}.

\bibitem[{Dosovitskiy et~al.(2021)Dosovitskiy, Beyer, Kolesnikov, Weissenborn, Zhai, Unterthiner, Dehghani, Minderer, Heigold, Gelly et~al.}]{book13}
Dosovitskiy, A.; Beyer, L.; Kolesnikov, A.; Weissenborn, D.; Zhai, X.; Unterthiner, T.; Dehghani, M.; Minderer, M.; Heigold, G.; Gelly, S.; et~al. 2021.
\newblock An image is worth 16x16 words: Transformers for image recognition at scale.
\newblock In \emph{International Conference on Learning Representations}.

\bibitem[{Fang et~al.(2025)Fang, Hsu, Lee, and Lee}]{book57}
Fang, Z.; Hsu, D.; Lee, G.~H.; and Lee, G.~H. 2025.
\newblock Neuralized Markov Random Field for Interaction-Aware Stochastic Human Trajectory Prediction.
\newblock In \emph{International Conference on Learning Representations}.

\bibitem[{Feichtenhofer et~al.(2022)Feichtenhofer, Li, He et~al.}]{book12}
Feichtenhofer, C.; Li, Y.; He, K.; et~al. 2022.
\newblock Masked autoencoders as spatiotemporal learners.
\newblock \emph{Advances in neural information processing systems}, 35: 35946--35958.

\bibitem[{Fu et~al.(2025)Fu, Yan, Wang, Li, and Liao}]{book62}
Fu, Y.; Yan, Q.; Wang, L.; Li, K.; and Liao, R. 2025.
\newblock Moflow: One-step flow matching for human trajectory forecasting via implicit maximum likelihood estimation based distillation.
\newblock In \emph{Proceedings of the IEEE/CVF Conference on Computer Vision and Pattern Recognition}, 17282--17293.

\bibitem[{Gao et~al.(2020)Gao, Sun, Zhao, Shen, Anguelov, Li, and Schmid}]{book8}
Gao, J.; Sun, C.; Zhao, H.; Shen, Y.; Anguelov, D.; Li, C.; and Schmid, C. 2020.
\newblock Vectornet: Encoding hd maps and agent dynamics from vectorized representation.
\newblock In \emph{Proceedings of the IEEE/CVF conference on computer vision and pattern recognition}, 11525--11533.

\bibitem[{Gu et~al.(2022)Gu, Chen, Li, Lin, Rao, Zhou, and Lu}]{book31}
Gu, T.; Chen, G.; Li, J.; Lin, C.; Rao, Y.; Zhou, J.; and Lu, J. 2022.
\newblock Stochastic trajectory prediction via motion indeterminacy diffusion.
\newblock In \emph{Proceedings of the IEEE/CVF Conference on Computer Vision and Pattern Recognition}, 17113--17122.

\bibitem[{Guo, Liu, and Pan(2022)}]{book9}
Guo, K.; Liu, W.; and Pan, J. 2022.
\newblock End-to-end trajectory distribution prediction based on occupancy grid maps.
\newblock In \emph{Proceedings of the IEEE/CVF Conference on Computer Vision and Pattern Recognition}, 2242--2251.

\bibitem[{Gupta et~al.(2018)Gupta, Johnson, Fei-Fei, Savarese, and Alahi}]{book29}
Gupta, A.; Johnson, J.; Fei-Fei, L.; Savarese, S.; and Alahi, A. 2018.
\newblock Social gan: Socially acceptable trajectories with generative adversarial networks.
\newblock In \emph{Proceedings of the IEEE/CVF Conference on Computer Vision and Pattern Recognition}, 2255--2264.

\bibitem[{Hochreiter and Schmidhuber(1997)}]{book17}
Hochreiter, S.; and Schmidhuber, J. 1997.
\newblock Long Short-Term Memory.
\newblock \emph{Neural Computation}, 9(8): 1735--1780.

\bibitem[{Huang et~al.(2019)Huang, Bi, Li, Mao, and Wang}]{book25}
Huang, Y.; Bi, H.; Li, Z.; Mao, T.; and Wang, Z. 2019.
\newblock Stgat: Modeling spatial-temporal interactions for human trajectory prediction.
\newblock In \emph{Proceedings of the IEEE/CVF International Conference on Computer Vision}, 6272--6281.

\bibitem[{Jacobs et~al.(1991)Jacobs, Jordan, Nowlan, and Hinton}]{book64}
Jacobs, R.~A.; Jordan, M.~I.; Nowlan, S.~J.; and Hinton, G.~E. 1991.
\newblock Adaptive mixtures of local experts.
\newblock \emph{Neural computation}, 3(1): 79--87.

\bibitem[{Jiang et~al.(2023)Jiang, Cornman, Park, Sapp, Zhou, Anguelov et~al.}]{book1}
Jiang, C.; Cornman, A.; Park, C.; Sapp, B.; Zhou, Y.; Anguelov, D.; et~al. 2023.
\newblock Motiondiffuser: Controllable multi-agent motion prediction using diffusion.
\newblock In \emph{Proceedings of the IEEE/CVF conference on computer vision and pattern recognition}, 9644--9653.

\bibitem[{Karnan et~al.(2022)Karnan, Nair, Xiao, Warnell, Pirk, Toshev, Hart, Biswas, and Stone}]{book3}
Karnan, H.; Nair, A.; Xiao, X.; Warnell, G.; Pirk, S.; Toshev, A.; Hart, J.; Biswas, J.; and Stone, P. 2022.
\newblock Socially compliant navigation dataset (scand): A large-scale dataset of demonstrations for social navigation.
\newblock \emph{IEEE Robotics and Automation Letters}, 7(4): 11807--11814.

\bibitem[{Lee et~al.(2024)Lee, Lee, Yu, Kim, and Lee}]{book37}
Lee, S.; Lee, J.; Yu, Y.; Kim, T.; and Lee, K. 2024.
\newblock MART: MultiscAle Relational Transformer Networks for Multi-agent Trajectory Prediction.
\newblock In \emph{European Conference on Computer Vision}, 89--107.

\bibitem[{Liu et~al.(2024{\natexlab{a}})Liu, Zhu, Yao, Mao, and Wang}]{book39}
Liu, S.; Zhu, Y.; Yao, P.; Mao, T.; and Wang, Z. 2024{\natexlab{a}}.
\newblock SpectrumNet: Spectrum-Based Trajectory Encode Neural Network for Pedestrian Trajectory Prediction.
\newblock In \emph{2024 IEEE International Conference on Acoustics, Speech and Signal Processing (ICASSP)}, 7075--7079.

\bibitem[{Liu et~al.(2024{\natexlab{b}})Liu, Dong, Lin, and Ye}]{book33}
Liu, Y.; Dong, X.; Lin, Y.; and Ye, M. 2024{\natexlab{b}}.
\newblock DifTraj: Diffusion Inspired by Intrinsic Intention and Extrinsic Interaction for Multi-Modal Trajectory Prediction.
\newblock In \emph{Proceedings of the International Joint Conference on Artificial Intelligence}, 1128--1136.

\bibitem[{Liu et~al.(2021)Liu, Lin, Cao, Hu, Wei, Zhang, Lin, and Guo}]{book11}
Liu, Z.; Lin, Y.; Cao, Y.; Hu, H.; Wei, Y.; Zhang, Z.; Lin, S.; and Guo, B. 2021.
\newblock Swin transformer: Hierarchical vision transformer using shifted windows.
\newblock In \emph{Proceedings of the IEEE/CVF international conference on computer vision}, 10012--10022.

\bibitem[{Loshchilov and Hutter(2019)}]{book51}
Loshchilov, I.; and Hutter, F. 2019.
\newblock Decoupled Weight Decay Regularization.
\newblock In \emph{International Conference on Learning Representations}.

\bibitem[{Mangalam et~al.(2020)Mangalam, Girase, Agarwal, Lee, Adeli, Malik, and Gaidon}]{book27}
Mangalam, K.; Girase, H.; Agarwal, S.; Lee, K.-H.; Adeli, E.; Malik, J.; and Gaidon, A. 2020.
\newblock It is Not the Journey but the Destination: Endpoint Conditioned Trajectory Prediction.
\newblock In \emph{European Conference on Computer Vision}, 759--776.

\bibitem[{Mao et~al.(2019)Mao, Liu, Salzmann, and Li}]{book53}
Mao, W.; Liu, M.; Salzmann, M.; and Li, H. 2019.
\newblock Learning trajectory dependencies for human motion prediction.
\newblock In \emph{Proceedings of the IEEE/CVF international conference on computer vision}, 9489--9497.

\bibitem[{Mao et~al.(2023)Mao, Xu, Zhu, Chen, and Wang}]{book32}
Mao, W.; Xu, C.; Zhu, Q.; Chen, S.; and Wang, Y. 2023.
\newblock Leapfrog diffusion model for stochastic trajectory prediction.
\newblock In \emph{Proceedings of the IEEE/CVF Conference on Computer Vision and Pattern Recognition}, 5517--5526.

\bibitem[{Martin-Martin et~al.(2021)Martin-Martin, Patel, Rezatofighi, Shenoi, Gwak, Frankel, Sadeghian, and Savarese}]{book49}
Martin-Martin, R.; Patel, M.; Rezatofighi, H.; Shenoi, A.; Gwak, J.; Frankel, E.; Sadeghian, A.; and Savarese, S. 2021.
\newblock Jrdb: A dataset and benchmark of egocentric robot visual perception of humans in built environments.
\newblock \emph{IEEE transactions on pattern analysis and machine intelligence}, 45(6): 6748--6765.

\bibitem[{Mehran, Oyama, and Shah(2009)}]{book21}
Mehran, R.; Oyama, A.; and Shah, M. 2009.
\newblock Abnormal crowd behavior detection using social force model.
\newblock In \emph{Proceedings of the IEEE/CVF Conference on Computer Vision and Pattern Recognition}, 935--942.

\bibitem[{Mohamed et~al.(2020)Mohamed, Qian, Elhoseiny, and Claudel}]{book26}
Mohamed, A.; Qian, K.; Elhoseiny, M.; and Claudel, C. 2020.
\newblock Social-stgcnn: A social spatio-temporal graph convolutional neural network for human trajectory prediction.
\newblock In \emph{Proceedings of the IEEE/CVF Conference on Computer Vision and Pattern Recognition}, 14424--14432.

\bibitem[{Morris and Trivedi(2011)}]{book22}
Morris, B.~T.; and Trivedi, M.~M. 2011.
\newblock Trajectory learning for activity understanding: Unsupervised, multilevel, and long-term adaptive approach.
\newblock \emph{IEEE Transactions on Pattern Analysis and Machine Intelligence}, 33(11): 2287--2301.

\bibitem[{Nie et~al.(2023)Nie, Nguyen, Sinthong, and Kalagnanam}]{book15}
Nie, Y.; Nguyen, N.~H.; Sinthong, P.; and Kalagnanam, J. 2023.
\newblock A Time Series is Worth 64 Words: Long-term Forecasting with Transformers.
\newblock In \emph{International Conference on Learning Representations}.

\bibitem[{Park, Jeong, and Yoon(2024)}]{book65}
Park, D.; Jeong, J.; and Yoon, K.-J. 2024.
\newblock Improving transferability for cross-domain trajectory prediction via neural stochastic differential equation.
\newblock In \emph{Proceedings of the AAAI Conference on Artificial Intelligence}, volume~38, 10145--10154.

\bibitem[{Paszke et~al.(2019)Paszke, Gross, Massa, Lerer, Bradbury, Chanan, Killeen, Lin, Gimelshein, Antiga et~al.}]{book50}
Paszke, A.; Gross, S.; Massa, F.; Lerer, A.; Bradbury, J.; Chanan, G.; Killeen, T.; Lin, Z.; Gimelshein, N.; Antiga, L.; et~al. 2019.
\newblock Pytorch: An imperative style, high-performance deep learning library.
\newblock In \emph{Advances in Neural Information Processing Systems}, volume~32.

\bibitem[{Pellegrini, Ess, and Gool(2010)}]{book46}
Pellegrini, S.; Ess, A.; and Gool, L.~V. 2010.
\newblock Improving Data Association by Joint Modeling of Pedestrian Trajectories and Groupings.
\newblock In \emph{European Conference on Computer Vision}, 452--465.

\bibitem[{Pellegrini et~al.(2009)Pellegrini, Ess, Schindler, and Van~Gool}]{book52}
Pellegrini, S.; Ess, A.; Schindler, K.; and Van~Gool, L. 2009.
\newblock You'll never walk alone: Modeling social behavior for multi-target tracking.
\newblock In \emph{Proceedings of the IEEE/CVF International Conference on Computer Vision}, 261--268.

\bibitem[{Phan-Minh et~al.(2020)Phan-Minh, Grigore, Boulton, Beijbom, and Wolff}]{book7}
Phan-Minh, T.; Grigore, E.~C.; Boulton, F.~A.; Beijbom, O.; and Wolff, E.~M. 2020.
\newblock Covernet: Multimodal behavior prediction using trajectory sets.
\newblock In \emph{Proceedings of the IEEE/CVF conference on computer vision and pattern recognition}, 14074--14083.

\bibitem[{Robicquet et~al.(2016)Robicquet, Sadeghian, Alahi, and Savarese}]{book47}
Robicquet, A.; Sadeghian, A.; Alahi, A.; and Savarese, S. 2016.
\newblock Learning Social Etiquette: Human Trajectory Understanding In Crowded Scenes.
\newblock In \emph{European Conference on Computer Vision}, 549--565.

\bibitem[{Ronneberger, Fischer, and Brox(2015)}]{book44}
Ronneberger, O.; Fischer, P.; and Brox, T. 2015.
\newblock U-net: Convolutional networks for biomedical image segmentation.
\newblock In \emph{Medical image computing and computer-assisted intervention--MICCAI 2015}, 234--241.

\bibitem[{Saadatnejad et~al.(2024)Saadatnejad, Gao, Messaoud, and Alahi}]{book56}
Saadatnejad, S.; Gao, Y.; Messaoud, K.; and Alahi, A. 2024.
\newblock Social-Transmotion: Promptable Human Trajectory Prediction.
\newblock In \emph{International Conference on Learning Representations}.

\bibitem[{Sadeghian et~al.(2019)Sadeghian, Kosaraju, Sadeghian, Hirose, Rezatofighi, and Savarese}]{book30}
Sadeghian, A.; Kosaraju, V.; Sadeghian, A.; Hirose, N.; Rezatofighi, H.; and Savarese, S. 2019.
\newblock Sophie: An attentive gan for predicting paths compliant to social and physical constraints.
\newblock In \emph{Proceedings of the IEEE/CVF Conference on Computer Vision and Pattern Recognition}, 1349--1358.

\bibitem[{Salzmann et~al.(2020)Salzmann, Ivanovic, Chakravarty, and Pavone}]{book6}
Salzmann, T.; Ivanovic, B.; Chakravarty, P.; and Pavone, M. 2020.
\newblock Trajectron++: Dynamically-feasible trajectory forecasting with heterogeneous data.
\newblock In \emph{European Conference on Computer Vision}, 683--700.

\bibitem[{Taketsugu et~al.(2025)Taketsugu, Oba, Maeda, Nobuhara, and Ukita}]{book63}
Taketsugu, H.; Oba, T.; Maeda, T.; Nobuhara, S.; and Ukita, N. 2025.
\newblock Physical Plausibility-aware Trajectory Prediction via Locomotion Embodiment.
\newblock In \emph{Proceedings of the IEEE/CVF Conference on Computer Vision and Pattern Recognition}.

\bibitem[{Tang and Zhang(2025)}]{book16}
Tang, P.; and Zhang, W. 2025.
\newblock Unlocking the Power of Patch: Patch-Based MLP for Long-Term Time Series Forecasting.
\newblock In \emph{Proceedings of the AAAI Conference on Artificial Intelligence}, volume~39, 12640--12648.

\bibitem[{Vaswani et~al.(2017)Vaswani, Shazeer, Parmar, Uszkoreit, Jones, Gomez, Kaiser, and Polosukhin}]{book34}
Vaswani, A.; Shazeer, N.; Parmar, N.; Uszkoreit, J.; Jones, L.; Gomez, A.~N.; Kaiser, {\L}.; and Polosukhin, I. 2017.
\newblock Attention is all you need.
\newblock \emph{Advances in Neural Information Processing Systems}, 30.

\bibitem[{Vemula, Muelling, and Oh(2018)}]{book24}
Vemula, A.; Muelling, K.; and Oh, J. 2018.
\newblock Social attention: Modeling attention in human crowds.
\newblock In \emph{2018 IEEE international Conference on Robotics and Automation (ICRA)}, 4601--4607.

\bibitem[{Weng et~al.(2023)Weng, Hoshino, Ramanan, and Kitani}]{book45}
Weng, E.; Hoshino, H.; Ramanan, D.; and Kitani, K. 2023.
\newblock Joint metrics matter: A better standard for trajectory forecasting.
\newblock In \emph{Proceedings of the IEEE/CVF International Conference on Computer Vision}, 20315--20326.

\bibitem[{Wong et~al.(2022)Wong, Xia, Hong, Peng, Yuan, Cao, Yang, and You}]{book18}
Wong, C.; Xia, B.; Hong, Z.; Peng, Q.; Yuan, W.; Cao, Q.; Yang, Y.; and You, X. 2022.
\newblock View vertically: A hierarchical network for trajectory prediction via fourier spectrums.
\newblock In \emph{European Conference on Computer Vision}, 682--700.

\bibitem[{Wong et~al.(2023)Wong, Xia, Peng, and You}]{book20}
Wong, C.; Xia, B.; Peng, Q.; and You, X. 2023.
\newblock Another vertical view: A hierarchical network for heterogeneous trajectory prediction via spectrums.
\newblock \emph{arXiv preprint arXiv:2304.05106}.

\bibitem[{Wong et~al.(2024)Wong, Xia, Zou, Wang, and You}]{book66}
Wong, C.; Xia, B.; Zou, Z.; Wang, Y.; and You, X. 2024.
\newblock SocialCircle: Learning the Angle-based Social Interaction Representation for Pedestrian Trajectory Prediction.
\newblock In \emph{Proceedings of the IEEE/CVF Conference on Computer Vision and Pattern Recognition}, 19005--19015.

\bibitem[{Wu et~al.(2023)Wu, Hu, Liu, Zhou, Wang, and Long}]{book40}
Wu, H.; Hu, T.; Liu, Y.; Zhou, H.; Wang, J.; and Long, M. 2023.
\newblock TimesNet: Temporal 2D-Variation Modeling for General Time Series Analysis.
\newblock In \emph{International Conference on Learning Representations}.

\bibitem[{Xu et~al.(2022{\natexlab{a}})Xu, Li, Ni, Zhang, and Chen}]{book28}
Xu, C.; Li, M.; Ni, Z.; Zhang, Y.; and Chen, S. 2022{\natexlab{a}}.
\newblock Groupnet: Multiscale hypergraph neural networks for trajectory prediction with relational reasoning.
\newblock In \emph{Proceedings of the IEEE/CVF Conference on Computer Vision and Pattern Recognition}, 6498--6507.

\bibitem[{Xu et~al.(2022{\natexlab{b}})Xu, Mao, Zhang, and Chen}]{book59}
Xu, C.; Mao, W.; Zhang, W.; and Chen, S. 2022{\natexlab{b}}.
\newblock Remember intentions: Retrospective-memory-based trajectory prediction.
\newblock In \emph{Proceedings of the IEEE/CVF Conference on Computer Vision and Pattern Recognition}, 6488--6497.

\bibitem[{Xu et~al.(2023)Xu, Tan, Tan, Chen, Wang, Wang, and Wang}]{book55}
Xu, C.; Tan, R.~T.; Tan, Y.; Chen, S.; Wang, Y.~G.; Wang, X.; and Wang, Y. 2023.
\newblock Eqmotion: Equivariant multi-agent motion prediction with invariant interaction reasoning.
\newblock In \emph{Proceedings of the IEEE/CVF conference on computer vision and pattern recognition}, 1410--1420.

\bibitem[{Yu et~al.(2020)Yu, Ma, Ren, Zhao, and Yi}]{book58}
Yu, C.; Ma, X.; Ren, J.; Zhao, H.; and Yi, S. 2020.
\newblock Spatio-temporal graph transformer networks for pedestrian trajectory prediction.
\newblock In \emph{Computer Vision--ECCV 2020: 16th European Conference, Glasgow, UK, August 23--28, 2020, Proceedings, Part XII 16}, 507--523.

\bibitem[{Yuan et~al.(2021)Yuan, Weng, Ou, and Kitani}]{book35}
Yuan, Y.; Weng, X.; Ou, Y.; and Kitani, K.~M. 2021.
\newblock Agentformer: Agent-aware transformers for socio-temporal multi-agent forecasting.
\newblock In \emph{Proceedings of the IEEE/CVF international conference on computer vision}, 9813--9823.

\bibitem[{Yue et~al.(2014)Yue, Lucey, Carr, Bialkowski, and Matthews}]{book48}
Yue, Y.; Lucey, P.; Carr, P.; Bialkowski, A.; and Matthews, I. 2014.
\newblock Learning Fine-Grained Spatial Models for Dynamic Sports Play Prediction.
\newblock In \emph{2014 IEEE International Conference on Data Mining}, 670--679.

\bibitem[{Zhang et~al.(2023)Zhang, Guo, Zhou, Zhang, and Lin}]{book19}
Zhang, Z.; Guo, D.; Zhou, S.; Zhang, J.; and Lin, Y. 2023.
\newblock Flight trajectory prediction enabled by time-frequency wavelet transform.
\newblock \emph{Nature Communications}, 14(1): 5258.

\bibitem[{Zhou et~al.(2021)Zhou, Zhang, Peng, Zhang, Li, Xiong, and Zhang}]{book42}
Zhou, H.; Zhang, S.; Peng, J.; Zhang, S.; Li, J.; Xiong, H.; and Zhang, W. 2021.
\newblock Informer: Beyond efficient transformer for long sequence time-series forecasting.
\newblock In \emph{Proceedings of the AAAI conference on artificial intelligence}, volume~35, 11106--11115.

\bibitem[{Zhou et~al.(2022{\natexlab{a}})Zhou, Ma, Wen, Sun, Yao, Yin, Jin et~al.}]{book43}
Zhou, T.; Ma, Z.; Wen, Q.; Sun, L.; Yao, T.; Yin, W.; Jin, R.; et~al. 2022{\natexlab{a}}.
\newblock Film: Frequency improved legendre memory model for long-term time series forecasting.
\newblock \emph{Advances in Neural Information Processing Systems}, 35: 12677--12690.

\bibitem[{Zhou et~al.(2022{\natexlab{b}})Zhou, Ma, Wen, Wang, Sun, and Jin}]{book41}
Zhou, T.; Ma, Z.; Wen, Q.; Wang, X.; Sun, L.; and Jin, R. 2022{\natexlab{b}}.
\newblock Fedformer: Frequency enhanced decomposed transformer for long-term series forecasting.
\newblock In \emph{International Conference on Machine Learning}, 27268--27286.

\bibitem[{Zhu et~al.(2021)Zhu, Claramunt, Brito, and Alonso-Mora}]{book4}
Zhu, H.; Claramunt, F.~M.; Brito, B.; and Alonso-Mora, J. 2021.
\newblock Learning interaction-aware trajectory predictions for decentralized multi-robot motion planning in dynamic environments.
\newblock \emph{IEEE Robotics and Automation Letters}, 6(2): 2256--2263.

\end{thebibliography}

\end{document}